\documentclass[letterpaper, 10 pt, journal, twoside]{IEEEtran}

\usepackage{graphics} %
\usepackage{epsfig} %
\usepackage{amsmath} %
\usepackage{amssymb}  %
\usepackage{tensor}
\usepackage[per-mode=symbol,binary-units,detect-all]{siunitx}
\newcommand{\norm}[1]{\|#1\|}

\usepackage{cite}

\usepackage{booktabs}
\usepackage{xcolor}
\usepackage[linesnumbered,ruled,vlined]{algorithm2e}
\usepackage{multirow}
\DeclareMathOperator*{\argmax}{arg\,max}
\DeclareMathOperator*{\argmin}{arg\,min}
\usepackage{hyperref}
\usepackage{microtype}
\usepackage{xprintlen}

\newcommand{\X}{\mathcal{X}}
\newcommand{\Z}{\mathcal{Z}}

\newcommand{\hide}[1]{}
\newcommand{\bdmath}{\begin{dmath}}
\newcommand{\edmath}{\end{dmath}}
\newcommand{\beq}{\begin{equation}}
\newcommand{\eeq}{\end{equation}}
\newcommand{\bdm}{\begin{displaymath}}
\newcommand{\edm}{\end{displaymath}}
\newcommand{\bea}{\begin{eqnarray}}
\newcommand{\eea}{\end{eqnarray}}
\newcommand{\beal}{\beq \begin{array}{ll}}
\newcommand{\eeal}{\end{array} \eeq}
\newcommand{\beas}{\begin{eqnarray*}}
\newcommand{\eeas}{\end{eqnarray*}}
\newcommand{\ea}{\end{array}}
\newcommand{\bit}{\begin{itemize}}
\newcommand{\eit}{\end{itemize}}
\newcommand{\ben}{\begin{enumerate}}
\newcommand{\een}{\end{enumerate}}

\newcommand{\SEthree}{\ensuremath{\mathrm{SE}(3)}\xspace}

\newcommand{\calC}{{\cal C}}

\newcommand{\calF}{{\cal F}}

\newcommand{\calI}{{\cal I}}

\newcommand{\calL}{{\cal L}}

\newcommand{\calX}{{\cal X}}
\newcommand{\calZ}{{\cal Z}}

\newcommand{\world}{\mathtt{{W}}}

\newcommand{\imu}{\mathtt{{I}}}
\newcommand{\base}{\mathtt{{B}}}

\newcommand{\Landmark}{\boldsymbol{m}}

\newcommand{\Figure}{Fig.~}

\newcommand{\Equation}{}

\newcommand{\ie}{{i.e.,~}}

\newcommand{\etalcite}[2]{#1~et~al.~\cite{#2}}
\newcommand{\State}{\boldsymbol{x}}
\newcommand{\tran}{\mathbf{p}}
\newcommand{\vel}{\mathbf{v}}
\newcommand{\bias}{\mathbf{b}}
\newcommand{\R}{\mathbf{R}}
\newcommand{\Identity}{\mathbf{I}}
\newcommand{\T}{\mathbf{T}}
\newcommand{\Real}{\mathbb{R}}
\newcommand{\SO}{\mathrm{SO}}
\newcommand{\SOthree}{\SO(3)}
\newcommand{\ba}{{\bias}^a} %
\newcommand{\bg}{{\bias}^g} %
\newcommand{\dba}{\dot{\bias}^a} %
\newcommand{\dbg}{\dot{\bias}^g} %
\newcommand{\hba}{\hat{\bias}^a} %
\newcommand{\hbg}{\hat{\bias}^g} %

\newcommand{\rotvel}{\boldsymbol\omega}
\newcommand{\tw}{\tilde{\rotvel}}

\newcommand{\acc}{\mathbf{a}}
\newcommand{\tacc}{\tilde{\acc}}
\newcommand{\Base}{\mathtt{{B}}}
\newcommand{\Camera}{\mathtt{C}}
\newcommand{\Lidar}{\mathtt{L}}
\newcommand{\World}{\mathtt{W}}
\newcommand{\Imu}{\mathtt{I}}
\newcommand{\gravity}{\tensor[_{\World}]{\mathbf{g}}{}}
\newcommand{\noise}{\boldsymbol\eta}
\newcommand{\etag}{\noise^g}
\newcommand{\etaa}{\noise^a}

\newcommand{\Cov}{\mathbf{\Sigma}}

\newcommand{\transpose}{\mathsf{T}}
\newcommand{\residual}{\mathbf{r}}

\title{\LARGE \bf
Deep IMU Bias Inference for Robust \\ Visual-Inertial
Odometry with Factor Graphs
}

\author{Russell Buchanan$^{1}$, Varun Agrawal$^{2}$, Marco Camurri$^{1}$,  Frank
Dellaert$^{2}$, Maurice Fallon$^{1}$%
\thanks{$^{1}$ Oxford Robotics Institute, University of Oxford, UK 
               \texttt{\{russell, mcamurri, mfallon\}@robots.ox.ac.uk}}%
\thanks{$^{2}$ Institute for Robotics and Intelligent Machines,
  College of Computing, Georgia Institute of Technology
   \texttt{\{varunagrawal, dellaert\}@cc.gatech.edquickly divergeu}}%
}

\begin{document}

\maketitle
\thispagestyle{empty}
\pagestyle{empty}

\begin{abstract}
Visual Inertial Odometry (VIO) is one of the most established state estimation methods for mobile platforms. However, when visual tracking fails, VIO algorithms quickly diverge due to rapid error accumulation during inertial data integration. This error is typically modeled as a combination of additive Gaussian noise and a slowly changing bias which evolves as a random walk. In this work, we propose to train a neural network to learn the true bias evolution. We implement and compare two common sequential deep learning architectures: LSTMs
and Transformers. Our approach follows from recent learning-based inertial
estimators, but, instead of learning a motion model, we target IMU bias
explicitly, which allows us to generalize to locomotion patterns unseen in
training. We show that our proposed method improves state estimation in
visually challenging situations across a wide range of motions by quadrupedal
robots, walking humans, and drones. Our experiments show an average 15\%
reduction in drift rate, with much larger reductions when there is total vision
failure. Importantly, we also demonstrate that models trained with one
locomotion pattern (human walking) can be applied to another (quadruped robot
trotting) without retraining.
\end{abstract}

\section{Introduction}

State estimation for lightweight, mobile systems is a fundamental problem in robotics. Visual Inertial Odometry (VIO) is a common solution due to the small size and low cost of cameras and IMUs. The main weak-point of VIO is that when visual feature tracking fails, only the Inertial Measurement Unit (IMU) can be used.

Inexpensive and miniaturized micro-electromechanical systems (MEMS) IMUs have become ubiquitous in robotics and pedestrian tracking~\cite{Wisth_2021, super-odometry,
Geneva2020ICRA, Foxlin2005}, however IMU-only state estimation is severely affected by drift. This drift is the result of the accumulation of various errors in IMU integration collectively modeled as a combination of additive zero-mean Gaussian noise and a slowly changing bias. As a result, estimation which relies purely on IMU measurement integration is not feasible for more than a few seconds due to explosive accumulation of drift.

VIO is effective so long as visual features are available because they constrain the drift of IMU integration~\cite{Forster2017} and estimate the biases. When visual tracking fails completely, the system relies purely on IMU integration. When there is error in visual tracking, the estimator may update the bias estimates to make sense of the data. In other words, the estimated bias is not necessarily related to the underlying physical process, but a quantity that minimizes the residuals. If there were another way to infer the IMU biases, these problems could be diminished.

\begin{figure}[t]
\centering
\includegraphics[height=4.2cm]{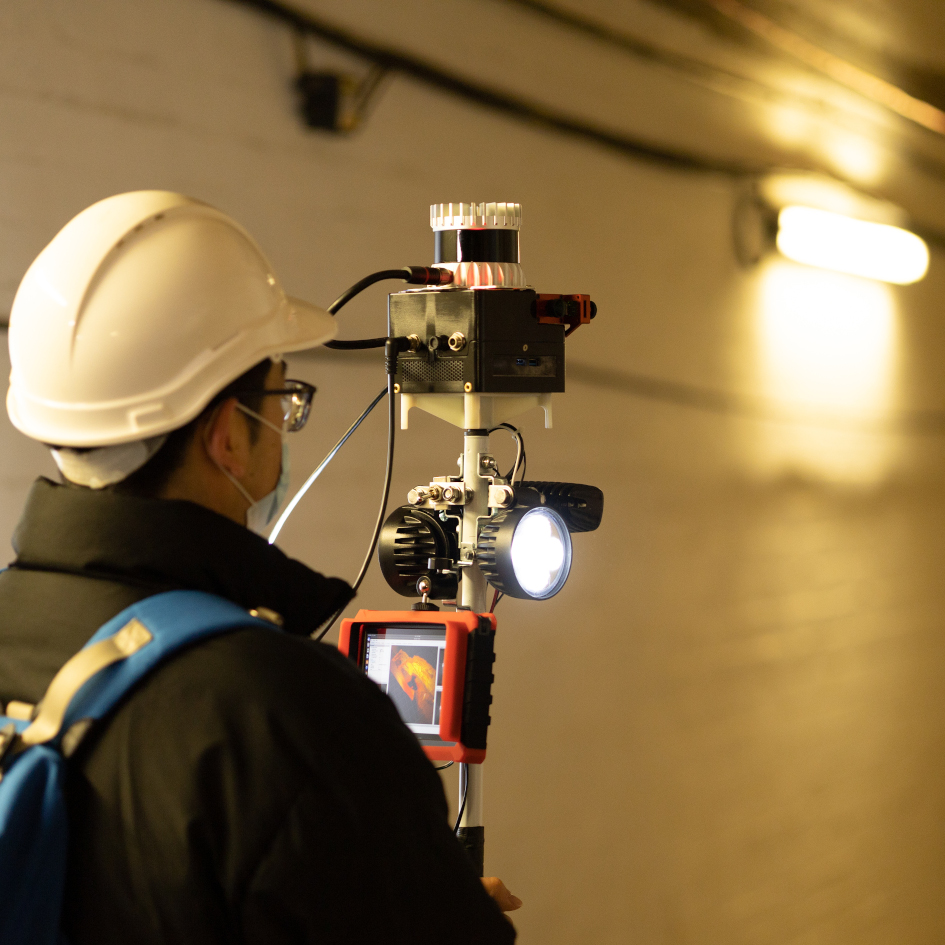}
\includegraphics[height=4.2cm]{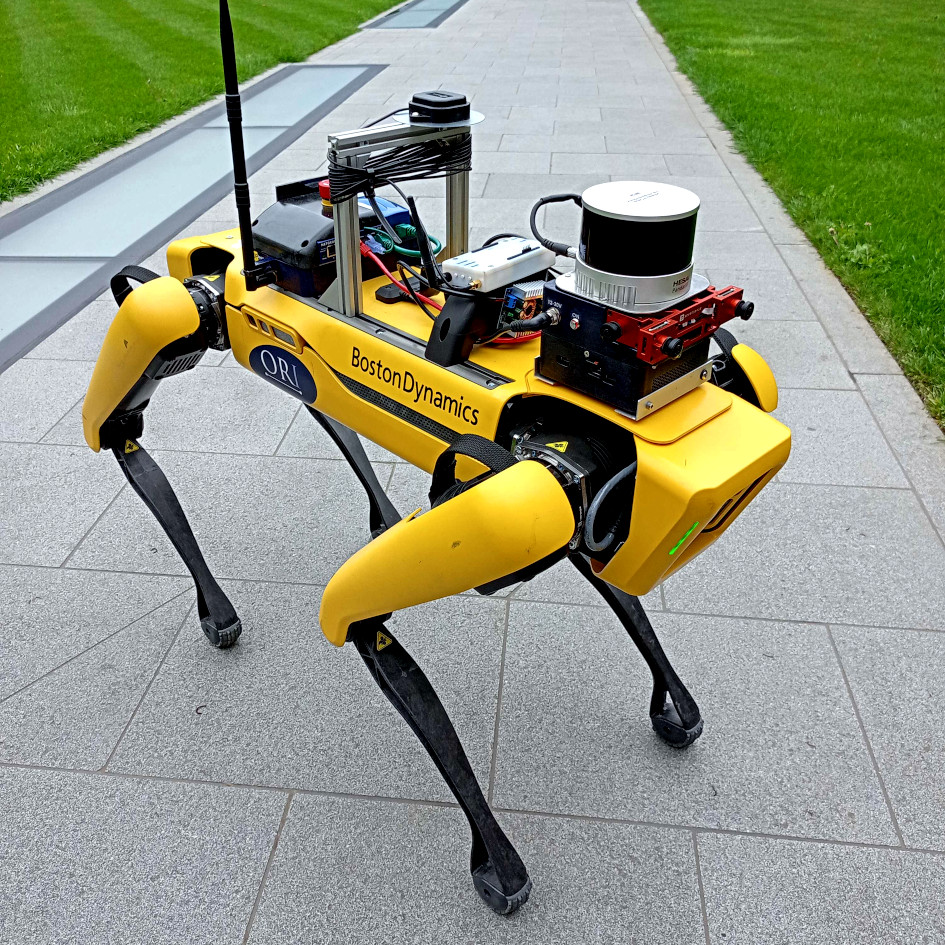}
\caption{Two of the platforms used in this work. \textbf{Left:} handheld device
including a SevenSense AlphaSense visual-inertial sensor. \textbf{Right:} the
same device mounted on a Boston Dynamics Spot quadruped. An IMU bias model from
a neural network trained on handheld data  was used to infer the biases on the Spot
dataset without any retraining --- despite the significantly different motion
pattern. The lidar was not used in these experiments.}
\label{fig:platforms}
\vspace{-5mm}
\end{figure}

Most VIO works have focused on improving visual tracking, however in this work we take a novel approach of \textit{improving IMU bias modeling}. We propose a new method which uses deep learning to estimate IMU biases directly from IMU measurements and past biases. We train a neural network to learn the evolution process of the biases of a specific IMU rather than assuming Brownian Motion. As a result, our method is \textit{device specific, not locomotion specific}, unlike similar IMU learning approaches, and does not require periodic motion. Our contributions can be listed as follows:
\begin{itemize}
\item A neural network capable of estimating IMU biases from a history of
measurements and biases. To the best of the authors' knowledge, this is the
first method capable of explicitly inferring the bias evolution of an IMU.
\item A performance comparison of two different network implementations (LSTM
and Transformer) and their integration as unary factors into a state estimator
based on factor graphs, for improved estimation in visually challenging
scenarios.
\item Real-world experiments on three different platforms with different types
of motions: pedestrian handheld, quadrupedal robot and drones. To the best of
the authors' knowledge, this is the first work that demonstrates an IMU learned
model trained on one locomotion modality and tested on another (handheld to
quadrupedal).
\end{itemize}

An additional minor contribution is the development of a ROS compatible,
open-source tool for calibrating an IMU using the Allan Variance
method\footnote{\url{https://github.com/ori-drs/allan_variance_ros}}.

\section{Related Work}
\label{sec:related-works}
In this section, we summarize the growing field of inertial learning from which our work follows. In Section~\ref{sec:learned-motion}, we discuss methods which learn motion
models from inertial data, while Section~\ref{sec:learned-noise} covers methods which
learn IMU noise models primarily for drone state estimation.

\subsection{Learning Inertial Motion Models}
\label{sec:learned-motion}
Recent works have trained neural networks with IMU data to learn motion models (typically of pedestrians) and to output estimates of velocity directly from IMU measurements.

The first data driven Inertial Navigation System (INS) was IoNet by
\etalcite{Chen}{Chen2018} which inferred 2D displacement and orientation change
from buffered IMU data. Later, \etalcite{Yan}{Yan2019} proposed RoNIN, a similar
network which inferred 2D velocity and orientation from raw IMU data.
\etalcite{Liu}{liu2020tlio} proposed TLIO, a method to infer 3D motion and
estimate high-fidelity trajectories in a filtering framework. They used an
Extended Kalman Filter (EKF) to estimate the full 6~DoF state. Process updates
were performed by traditional IMU mechanization while measurement updates from
the network were used as relative position measurements. This allowed the filter
to implicitly estimate IMU biases.

\begin{figure}[t]
\centering
\vspace{2mm}
\includegraphics[width=\columnwidth]{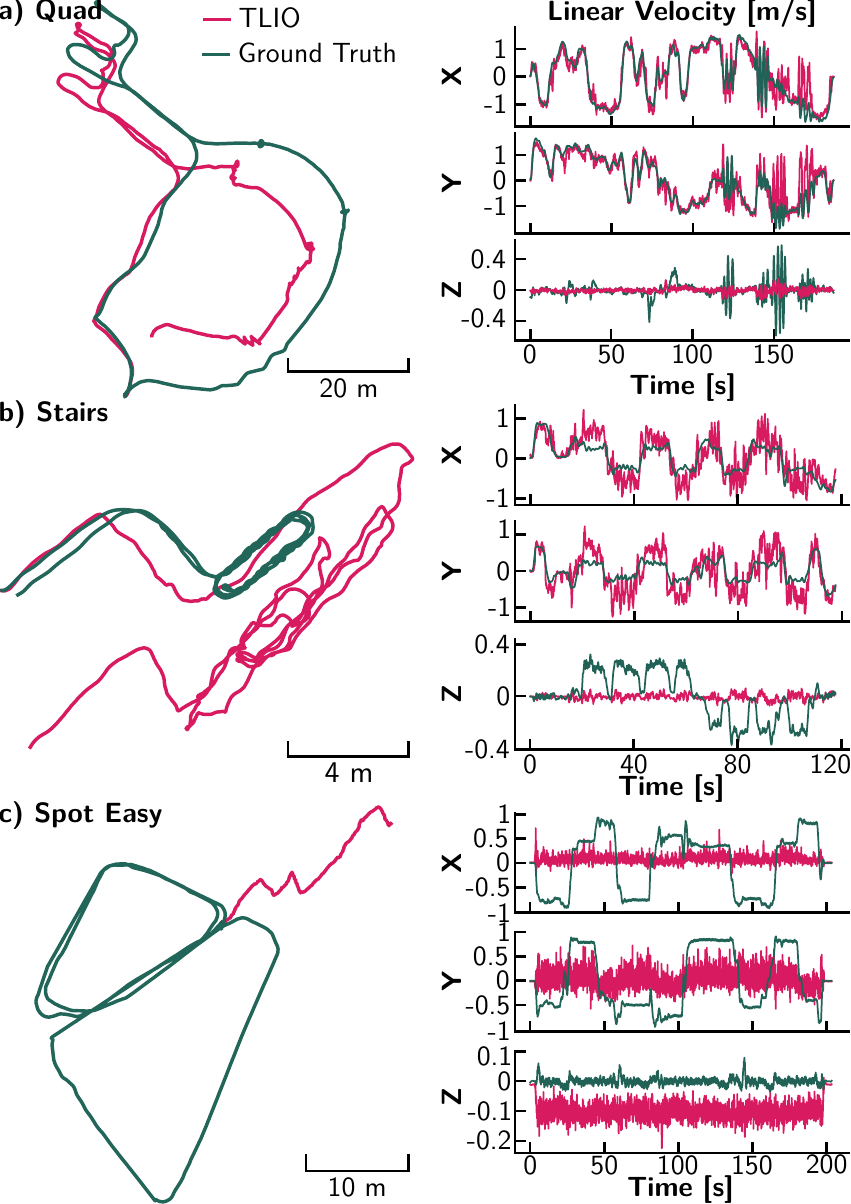}
\caption{\textbf{Top:} TLIO \cite{liu2020tlio} trained on handheld data on flat
ground, tested on flat ground. \textbf{Middle:} same network tested on a sequence
in which the person walks up and down stairs. Because stair climbing was not
present in the training set, inference fails significantly (Note $z$ velocity
estimates). \textbf{Bottom:} when applied to a quadrupedal robot,
because of the completely different locomotion modality, TLIO is unable to
estimate any position.}
\label{fig:motivation}
\vspace{-5mm}
\end{figure}

In our prior work, we adapted the approach of \cite{liu2020tlio} to incorporate
additional sensors such as cameras, lidar or legged robot kinematics in a
factor graph~\cite{buchanan-corl}. However, this method, like the ones discussed
above, was based on learning a motion model and was therefore susceptible to
failure if applied outside of the training domain. For example, in
Fig.~\ref{fig:motivation} we show results of TLIO which was trained with
handheld walking data on flat ground. As shown in Fig.~\ref{fig:motivation}~a), the
velocity estimates are reasonable for IMU-only odometry. In Fig.~\ref{fig:motivation}~b), we applied the same network to stair climbing, which was not present in the training
set and as a result, the $z$ velocity estimation fails completely. In Fig.~\ref{fig:motivation}~c)
plot we used the network with a quadruped and the model fails completely. This motivates the need for methods which are more IMU specific rather than
locomotion specific.

\subsection{Learning IMU Noise Models}
\label{sec:learned-noise}
A different approach to velocity estimation is learning IMU noise models. In this case, the IMU measurements are passed through a network trained to ``denoise'' them and output an estimate of the perfect IMU measurements. These denoised signals can then be used directly as input to a VIO pipeline.

\etalcite{Brossard}{Brossard20ral} used a Convolutional Neural Network (CNN) to
denoise a gyroscope using the output to correct for the true angular
velocity. The CNN used dilated convolutions to increase the temporal coverage over longer sequences. \etalcite{Zhang}{zhang2021imu} trained a Recurrent Neural Network (RNN) to
denoise both gyroscope and accelerometer measurements. Their network used IMU
measurements as input to estimate a corrected measurement which, when
integrated, reduced pose error. This eliminated the need for the model to learn the
underlying mechanization equations. Similarly, \etalcite{Steinbrener}{Steinbrener2022} performed denoising on IMU measurements, comparing LSTM and Transformer architectures, finding the LSTM to be more effective.

The main drawback of these methods is that they do not distinguish between
different noise sources and as a result, it is not clear what the network has learned to
remove from the IMU measurements. For example, most results were primarily demonstrated on
drones which, when flying, introduce high frequency vibrations affecting the IMU
measurements. It is unclear if the noise characteristics being
estimated was caused by the vibrations, the internal state of the IMU (i.e. the true bias), the motion of the drone, or a combination of the above. By explicitly  modeling
the IMU bias, in this paper we seek a motion independent method with more explainable
outputs.

\section{Problem Statement}
\label{sec:problem-statement}
\begin{figure}
\centering
\vspace{2mm}
\includegraphics[width=0.65\columnwidth]{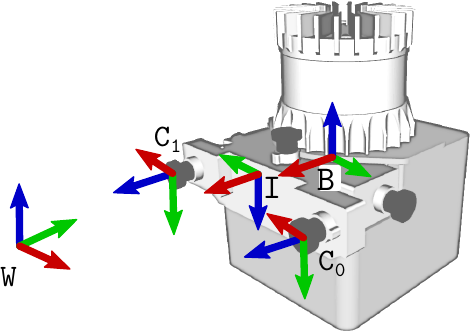}
\caption{Reference frames convention for the handheld device. The world frame
$\World$ is a fixed frame, while the base $\Base$, camera optical $\Camera$, IMU
$\Imu$, and lidar $\Lidar$ frames are attached to the moving device.}
\label{fig:coordinate-frames}
\vspace{-5mm}
\end{figure}

We aim to estimate the trajectory of a
mobile platform equipped with an
IMU and a stereo camera using sliding window based smoothing. The relevant reference frames of the platform are
shown in Fig.~\ref{fig:coordinate-frames} and includes an Earth-fixed world
frame $\World$, the platform-fixed base frame $\Base$, left and right camera frames $\Camera_0,
\Camera_1$, and IMU sensor frame $\Imu$.

\subsection{Definition of State Vector and Measurements}
\noindent The state of the platform at time $t_i$ is defined as:
\begin{equation}
\State_i \triangleq \left[\R_i,\tran_i,\vel_i, \bias^{g}_i \,\; \bias^{a}_i
\right] \in \SOthree \times \Real^{12}
\end{equation}
where: $\R_i = \R_{\world\base}(t_i)$ is the orientation of $\Base$ with respect to $\World$, $\tran_i = \tensor[_\world]{\tran}{_{\world\base}}(t_i)$ is the position,
$\vel_i = \tensor[_\base]{\vel}{_{\world\base}}(t_i)$ is the linear
velocity, and $\bias^{g}_i = \tensor[_\imu]{\bias}{^g}(t_i),\;
\bias^{a}_i = \tensor[_\imu]{\bias}{^a}(t_i)$ are the IMU gyroscope
and accelerometer biases, respectively.

We indicate the set of all states in the window as $\X_k = \{\State_i\}_{i \in
\mathsf{K}_k}$ where $\mathsf{K}_k$ are all the keyframe indices up to $t_k$.
Similarly, the measurements within the window are $\Z_k = \{\calI_{ij},
\calC_i\}_{i,j} \in \mathsf{K}_k$, where $\calI_{ij}$ are the IMU measurements
between two camera keyframes (with $i = j-1$), while $\calC_i$ include the
stereo image
pairs at time $t_i$.

\subsection{Maximum-A-Posteriori (MAP) Estimation}
We maximize the likelihood  of the measurements $\calZ_k$,
given the history of states $\calX_k$,
\begin{equation}
 \X^*_k = \argmax_{\X_k} p(\X_k|\Z_k) \propto
p(\X_0)p(\Z_k|\X_k)
\label{eq:posterior}
\end{equation}
The measurements are formulated as conditionally independent and
corrupted by white Gaussian noise. Therefore, \Equation
\eqref{eq:posterior}
can be expressed as the following least squares minimization:
\begin{multline}
\X^{*}_k = \argmin_{\X_k} \|\mathbf{r}_0\|^2_{\Sigma_0} + \sum_{i \in
\mathsf{K}_k}   \Big(
\|\mathbf{r}_{\calI_{ij}}\|^2_{\Sigma_{\calI_{ij}}}
+ \\ \sum_{\ell \in \mathsf{M}_i} \|\mathbf{r}_{\State_i,\Landmark_{\ell}}
\|^2_{\Sigma_{\State_i, \Landmark_\ell}} + \|\residual_{\ba_j}\|^2_{\hat
{\Cov}_{\ba}} +
\|\residual_{\bg_j}\|^2_{\hat{\Cov}_{\bg}} \Big)
\label{eq:cost-function}
\end{multline}
where each term is the residual associated to a measurement type, weighted by
the inverse of its covariance matrix, and will be detailed in Section
\ref{sec:state-estimator}.

\begin{figure}[t]
\centering
\vspace{2mm}
\includegraphics{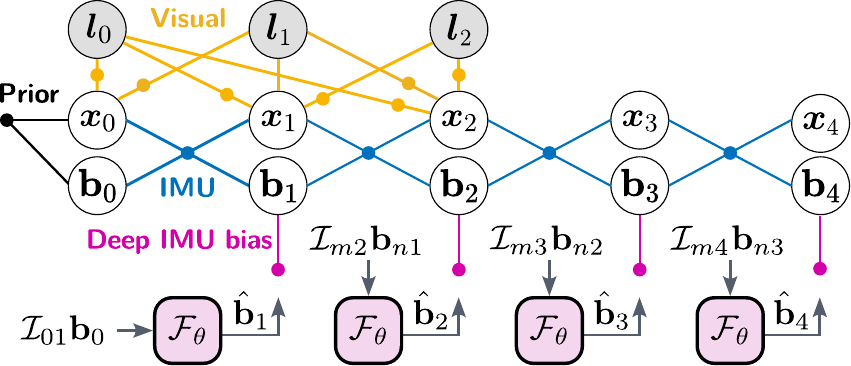}
\caption{Proposed factor graph framework with learned IMU estimates. A neural
network $\mathcal{F}_\theta$ trained on ground truth IMU biases takes a window of
past IMU measurements $\calI_{mj}$ and previous bias estimates $\bias_{ni}$ from
the optimizer and outputs the current bias estimate $\hat{\bias}_j$. This is
integrated into the factor graph as unary factor (magenta). This helps where
visual features cannot be tracked reliably (right side of figure where
landmarks are absent).
}
\label{fig:fg}
\vspace{-3mm}
\end{figure}

\section{Factor Graph Formulation}
\label{sec:state-estimator}
A factor graph can be used to graphically represent \Equation \eqref{eq:cost-function} as in \Figure\ref{fig:fg}. With slight abuse of notation, the IMU biases are shown separately from the state nodes $\State_i$ to highlight our contribution. In addition to the prior
factors (black), the graph includes: IMU preintegration (blue), visual
tracking (yellow) and our novel deep IMU bias (purple) factors. For
convenience, the first two are briefly reported in this section, while the last
one is detailed in Section~\ref{sec:bias-factor}.

\subsection{Preintegrated IMU Factors}
\label{sec:imu-factors}
We follow the standard manner of IMU measurement
integration from \cite{Forster2017} to constrain the pose, velocity, and
biases between two consecutive nodes of the graph. The residual has the following form:
\begin{equation}
\mathbf{r}_{\calI_{ij}}  = \left[ \mathbf{r}^\transpose_{\Delta
\R_{ij}}, \mathbf{r}^\transpose_{\Delta \vel_{ij}},
\mathbf{r}^\transpose_{\Delta \tran_{ij}},
\mathbf{r}_{\bias^a_{ij}},
\mathbf{r}_{\bias^g_{ij}} \right]
\end{equation}
For a detailed definition of the above residuals, see \cite{Forster2017}.

\subsection{Stereo and Mono Landmark Factors}
\label{sec:visual-factors}
A visual landmark in
Euclidean space $\Landmark_\ell \in
\Real^3$ is projected onto the image plane by the function $\pi: \SEthree
\times \Real^3 \mapsto \Real^2$, given the platform pose
$\T_i = \{\tran_i,\R_i\} \in \SEthree$. Given a landmark
$\Landmark_\ell$ and its coordinates $(u_{\ell}, v_{\ell}) \in
\Real^2$ on the image plane, the residual at state is computed
as:
\begin{equation}
	\mathbf{r}_{\State_i, \Landmark_\ell} =
	\left( \begin{array}{c}
		\pi_u^L(\T_i, \Landmark_\ell) - u^L_{i,\ell} \\
		\pi_u^R(\T_i, \Landmark_\ell) - u^R_{i,\ell} \\
		\pi_v(\T_i, \Landmark_\ell) - v_{i,\ell}
	\end{array} \right) \label{eq:stereo-residual}
\end{equation}
where $(u^{L}, v), (u^{R}, v)$ are the pixel locations of the landmark.
$\Sigma_{\Landmark}$ is computed using an uncertainty of \SI{0.25}{pixels} and the Dynamic Covariance Scaling robust loss function as in \cite{Wisth2022tro}.

\section{Deep IMU Bias Factors}
\label{sec:bias-factor}
A MEMS IMU measures its proper acceleration (e.g., equal to Earth's
gravity at rest) $\tacc \in \Real^3$ and rotational rate $\tw \in
\Real^3$. The absolute linear acceleration $\acc$ (e.g., null at rest) and the rotation
rate $\rotvel$ of a body expressed in an Earth-fixed coordinate frame $\World$
can be recovered as follows:
\begin{align}
\begin{split}
\tacc(t) =& \R_{\world\imu}^\transpose(t)(\acc(t) - \gravity) + \ba(t) +
\etaa(t) \\
\tw(t) =& \rotvel(t) + \bg(t) + \etag(t)
\end{split}
\end{align}

Where $\R_{\world\imu} \in \SOthree$ is the absolute orientation of the IMU and
$\gravity$ is the acceleration due to gravity expressed in $\World$. The
quantities $\etaa$ and $\etag$ represent additive noise present in all IMUs
and is modeled as zero-mean Gaussian distributions.

\subsection{Bias Noise Model}
The IMU biases $\ba$ and $\bg$ are due to physical properties of the IMU.
They change with each power cycle and continue to change slowly during
operation~\cite{strapdown-ins}.
Typically, their
evolution is modeled as a Brownian noise process, whose derivatives are sampled from zero-mean Gaussian distributions:
\begin{equation}
\label{eq:bias-process-continuous}
	    \dba(t) = \noise^{ba} \qquad
	    \dbg(t) = \noise^{bg}
\end{equation}

This can be re-written in discrete time by integrating between two time steps
$[t_i, t_j]$:
\begin{equation}
\ba_j = \ba_i + \noise^{bad} \qquad
\bg_j = \bg_i +  \noise^{bgd}
\label{eq:bias-process-discrete}
\end{equation}

Where $\noise^{bad}$ and $\noise^{bgd}$ are discrete zero mean Gaussian
distributions with
covariance $\mathbf{\Sigma}^{bad}$ and $\mathbf{\Sigma}^{bgd}$
respectively~\cite{Forster2017}.

\subsection{Learning the Bias Process Model}
\label{sec:bias-learning}

The Brownian noise model for IMU biases in
\Equation\eqref{eq:bias-process-discrete} is an approximation which does not
hold for
long periods of time. In practice, the true underlying
dynamics are a highly nonlinear function depending on vibrations,
impacts, and physical properties of the device \cite{strapdown-ins}. Thus,
modeling the IMU dynamics is
a problem that lends
itself to deep learning as it allows us to approximate highly nonlinear
functions. To this
end, instead of \Equation\eqref{eq:bias-process-discrete}, we propose a
deep
neural network $\calF$ with parameters $\theta$ defined as follows:
\begin{equation}
\label{eq:network}
\calF_\theta  : \left(\calI_{mj}, \ba_{ni}, \bg_{ni} \right) \mapsto (\hba_j,
\hbg_j)
\end{equation}
where the inputs are a buffer of IMU measurements $\calI_{mj}$ between
times $t_m$ and $t_j$, and the previous bias values $\ba_{ni}, \bg_{ni}$ between
$t_n$ and $t_i$.
The network outputs $(\hba_j, \hbg_j)$ are the estimates of the
IMU bias value at time $t_j$. Without loss of generality, in this section we assume the output of the network is generated at the camera keyframe rate.

\subsection{Deep IMU Bias Factors}
We incorporate the bias estimates from the network into a factor
graph-based state estimator. The estimates are modeled as unary factors on
the bias state, as shown in
Fig.~\ref{fig:fg}. The last two residuals of \Equation(\ref{eq:cost-function}),
 $\residual_{\ba_j}$ and $\residual_{\bg_j}$ correspond to:
\begin{equation}
\label{eq:residual}
	\residual_{\ba_j} =  \ba_j -
\hba_j \quad\quad
	\residual_{\bg_j} =  \bg_j -
\hbg_j
\end{equation}

The covariances $\hat{\Cov}_{\ba}$ and $\hat{\Cov}_{\bg}$ were tuned to fixed
values for our experiments, with  $\hat{\Cov}_{\ba} =
\Identity_3 \cdot 2.5\mathrm{e}^{-3} $ and $\hat{\Cov}_{\bg} =
\Identity_3 \cdot  2.5\mathrm{e}^{-5}$.
In future work, we intend to train the networks to provide
measurement uncertainty estimates as in~\cite{liu2020tlio,buchanan-corl}.

\begin{figure}[t]
\centering
\vspace{2mm}
\includegraphics[width=\columnwidth]{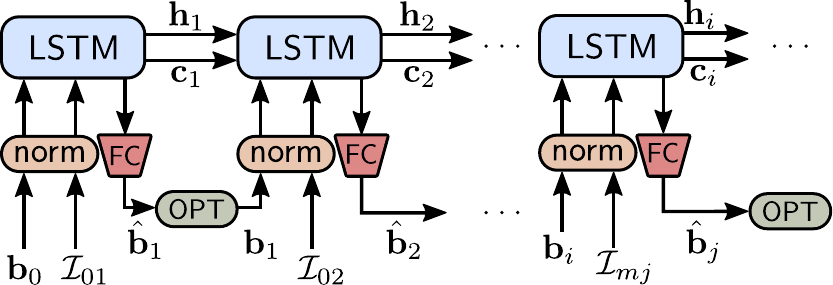}
\caption{LSTM architecture. A IMU data window $\mathcal{I}_{mj}$ of size $w$
(with $m = j - w$) and the previous bias $\bias_{i}$ are first normalized then
passed to the LSTM. The hidden state is preserved for the next
inference step and the output is passed through a fully connected layer to
predict a bias.}
\label{fig:lstm-model}
\end{figure}
\begin{figure}[t]
\includegraphics[width=\columnwidth]{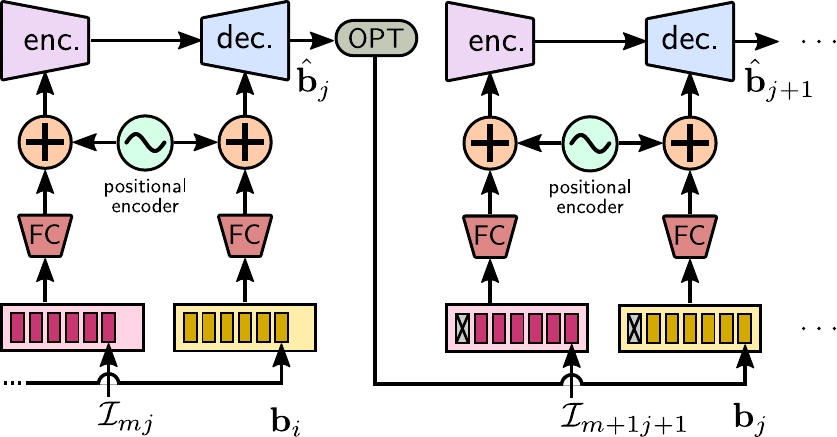}
\caption{Transformer architecture. A history of
$l$ IMU windows $\mathcal{I}_{mj}$ and biases $\bias_{i}$ are concatenated
and summed with a positional
encoding
before being passed to the Transformer.}
\label{fig:model}
\vspace{-5mm}
\end{figure}

\section{Network Architecture}
Since our learning task involves sequential data, two common network
architectures are well suited when implementing
\Equation\eqref{eq:network}. The first one is
based on Long Short-Term Memory (LSTM)~\cite{hochreiter1997long} which is
lightweight (see Fig.~\ref{fig:lstm-model}) and the second is based on
Transformers~\cite{transformers} which have been shown to have better
performance for longer-term sequences
(see \Figure\ref{fig:model}).

\subsection{LSTM}
The input to the LSTM is a single window of size $w$ of IMU measurements $\calI_{mj}$
(with $m = j - w$) and the previous bias estimate $\mathbf{b}_i$  (\ie $t_n = t_i$), which comes from the factor graph
optimizer. This exposes the network to bias estimates resulting from the fusion of additional sensors. These are normalized then passed to the LSTM with states
$\mathbf{h}_i$ and $\mathbf{c}_i$ which are preserved for the next bias
estimate. In this way, the LSTM $w$ batches of IMU measurements while
the memory can
observe the bias evolution over time. One limitation to this method is that,
over a long trajectory, the LSTM will eventually forget information from old
inputs.

\subsection{Transformer} The Transformer input is a history of
$l$ windows of IMU measurements and biases (\ie $m = j- l\cdot w, n = l$). Similar to the LSTM, biases added to the history come from
the estimator's optimizer. A history of information allows the
Transformer attention mechanism to recall older information.

\subsection{Loss Function}
\noindent We use the Mean Square Error (MSE) as a loss function:
\begin{equation}
\calL(\bias,\hat{\bias}) = \frac{1}{n}\sum^{n}_{k=1} \norm{\bias_k -
\hat{\bias}_k}^2
\label{eq:bias-loss}
\end{equation}
where two separate instances of the network are trained with
\eqref{eq:bias-loss} for accelerometer and gyroscope biases,
respectively.

\section{Network Implementation}
\subsection{Datasets}
The datasets used for training and experiments are listed in
Table~\ref{tab:sensors}. The first is the Newer College Multi-Camera Dataset
({NCD-Multi})~\cite{zhang2021multicamera} which uses an AlphaSense
inertial and multi-camera sensor (see Fig.~\ref{fig:platforms}), although we
only use the front facing stereo cameras. NCD-Multi was collected by a human
operator walking while carrying the device inside New College, Oxford. An
additional non-public sequence was collected using the same device in a
limestone mine~\cite{zhang2022ral}.

The second dataset, {Spot}, was recorded to
demonstrate generalization to motions patterns unseen in training. We placed the same sensor payload used in NCD-Multi on a quadruped robot (see Fig.~\ref{fig:platforms}
Right) and recorded two sequences only for testing.

Finally, we present results with the public {EuRoC} dataset~\cite{Burri25012016}. We included this dataset to show results alongside similar methods and to demonstrate application to a very different platform - a drone.

\begin{table}[t]
\centering
\vspace{2mm}
\caption{Datasets Used for Experiments.}
\label{tab:sensors}
\resizebox{\columnwidth}{!}{
\begin{tabular}{l|ccccc}
\toprule
\textbf{Dataset} & \textbf{Duration [s]} & \textbf{Length [m]} & \textbf{IMU} &
\textbf{Locom.}\\
\midrule
NCD-Multi &  3163 & 4512 & BMI085* & Handheld\\
Spot &  515 & 399 & BMI085* & Quadruped\\
EuRoC &  1349 & 893 & ADIS16448 & Drone\\
\bottomrule
\end{tabular}
}\\
\vspace{1mm}
\parbox[t]{\textwidth}{\footnotesize *Same IMU device.}
\vspace{-7mm}
\end{table}

\subsection{Training}
We trained one LSTM and one Transformer each for the public datasets (NCD-Multi
and EuRoC), for a total of four models. For each dataset, we selected hard sequences for testing (see \ref{sec:handheld}) and the remaining were split 75:25 for training and validation. The Spot dataset was only used for testing because, as detailed in
Section~\ref{sec:visual-challenges}, we demonstrate generalization to different
locomotion modalities by training a model on NCD-Multi and testing it on the Spot dataset.

For training, we used teacher forcing~\cite{teacherForcing}, providing the
network with ground truth biases. The ground truth biases for NCD-Multi were estimated using the lidar and a high resolution 3D model of the environment while the EuRoC dataset provides IMU bias estimates. We added noise to the ground truth biases which was sampled from a zero-mean Gaussian distribution with
covariances $\mathbf{\Sigma}^{bad}$ and $\mathbf{\Sigma}^{bgd}$ estimated from
Allan Variance analysis.

As in~\cite{liu2020tlio}, we rotate the IMU measurement windows
$\calI_{mj}$ into a gravity aligned frame. This prevents the effect of gravity
from significantly changing the apparent accelerometer biases.

We used the Adam optimizer with a learning rate of $10^{-5}$ and batch size
$32$. Training lasted 200 epochs for both LSTM and Transformer with the model
minimizing validation error used. Both networks take $\SI{\sim 2}{\hour}$ to train on a desktop computer with one Nvidia Titan X with \SI{12}{\giga\byte} of memory.

\subsection{Networks Models}
\subsubsection{LSTM} We used a 2-layer single direction LSTM with hidden state
size of $256$. We found that a larger network provided no improvement in
performance while a lighter-weight LSTM is capable of running online with
several inferences per second on a standard laptop with low-grade GPU. After
testing different options, we found that \SI{1}{\second} of IMU data (\ie $w =
10$) and an inference rate of \SI{2}{\hertz} (50\% data
overlap between consecutive inferences) to give the best performance and used
these settings for all experiments.

\subsubsection{Transformer}  We used an 8 headed Transformer with 2 encoder and
decoder layers and an embedding size of 512. Similar to LSTM, we used
\SI{1}{\second} IMU windows with a maximum history size of $l = 100$. This was
based on
analysis of the Allan Variance plots which found the IMU bias instability to
dominate noise generally around \SI{100}{\second} sampling times. Inference
was kept at \SI{1}{\hertz} since using a history of IMU windows makes
overlapping of input data unnecessary.

\begin{figure}[t]
\centering
\vspace{2mm}
\includegraphics[width=0.45\columnwidth]{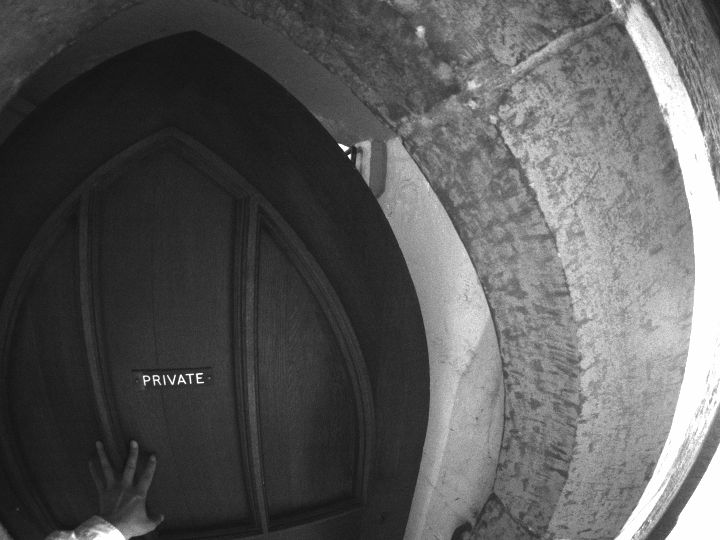}
\includegraphics[width=0.45\columnwidth]{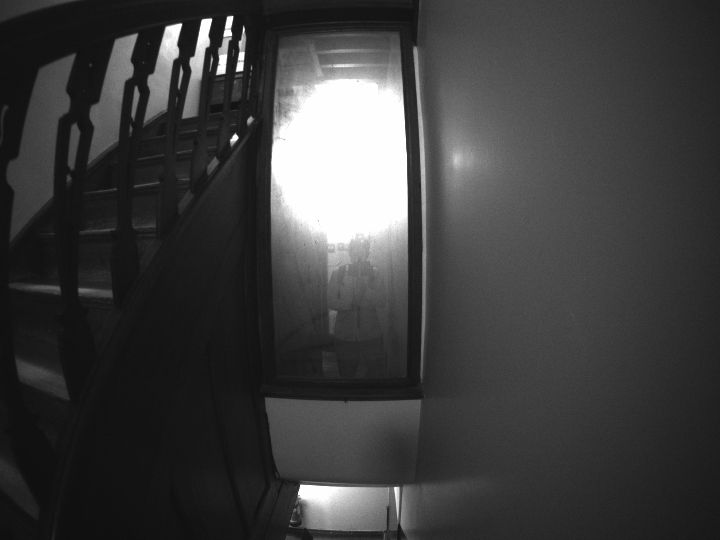}\\
\vspace{2mm}
\includegraphics[width=0.45\columnwidth]{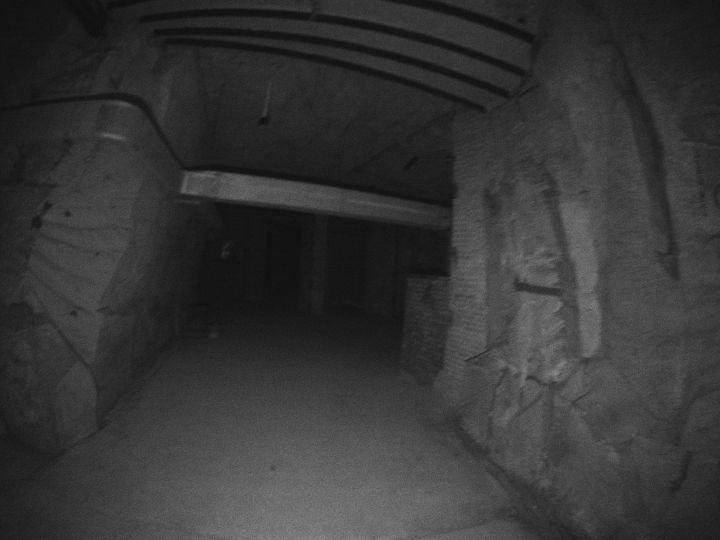}
\includegraphics[width=0.45\columnwidth]{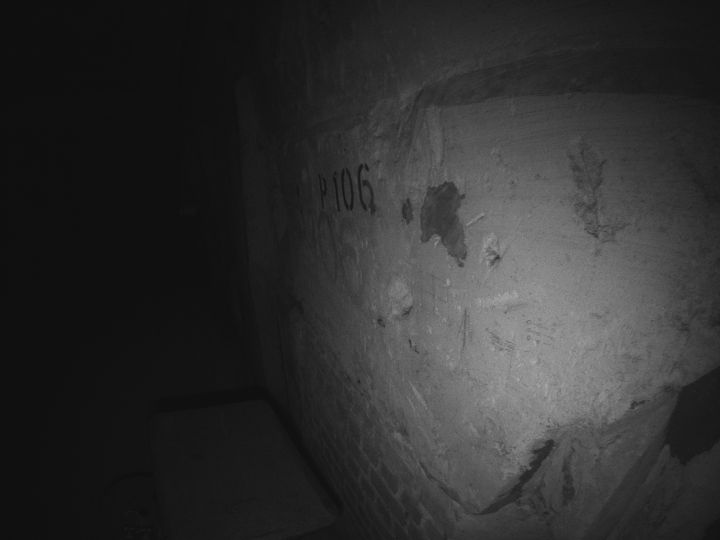}
\caption{Visual challenges in the NCD-Multi Dataset. \textbf{Top Row:} Images
from NCD-Multi Stairs sequence with door opening obscuring the cameras shown on
the left and tight spaces of about 1.\SI{5}{\meter} wide and a mirror shown on
the right. \textbf{Bottom Row:} Images from NCD-Multi Mine sequence with
darkness and the challenge of camera being held up close to a wall.}
\label{fig:challenges}
\vspace{-5mm}
\end{figure}

\section{Experimental Results}
\label{sec:visual-challenges}
In this section, we describe the experiments and results of the
proposed state estimator. We report relative
position and orientation errors as described in~\cite{strumBenchmark}. We
compare our method which uses bias predictions as in Fig.~\ref{fig:fg} to our
baseline method, which consists of the same factor graph but without the unary
factor described in Section \ref{sec:state-estimator}.

\subsection{Handheld Experiments}
\label{sec:handheld}
There are many common situations or environments in which visual odometry
systems degrade or even fully fail such as door-opening, narrow spaces or fast
rotations. We selected several sequences from NCD-Multi which exhibit these
challenges to test our method. These sequences were only used for testing and
contained situations unseen in training.

\begin{table}[t]
\centering
\vspace{2mm}
\caption{\SI{10}{\meter} Relative Translation/Rotation Error
[m]/[\si{\degree}]}
\resizebox{\columnwidth}{!}{
    \begin{tabular}{l|ccc}  
    \toprule
    \textbf{Experiment} & \textbf{VIO baseline} & \textbf{LSTM} &
\textbf{Transformer}\\
    \midrule
    NCD-Multi Stairs* & 0.23 / 3.22 & 0.19 / 2.88                   & %
\textbf{0.17} / %
\textbf{2.70}\\
    NCD-Multi Mine    & 0.46 / 2.59 & \textbf{0.33} / \textbf{1.71}  & 0.42 /
2.26 \\
    NCD-Multi Quad    & 0.35 / 1.38 & 0.28 / 1.36 & \textbf{0.25} /
\textbf{1.29}\\
    Spot Easy         & 0.31 / 1.21 & \textbf{0.27} / 1.02  & 0.31 /
\textbf{0.88} \\
    Spot Hard         & 0.38 / 1.57 & 0.34 /\textbf{1.42}          &
\textbf{0.30} / \textbf{1.42}\\
    \bottomrule
    \end{tabular}
    }
    
    \vspace{1mm}
    \parbox[t]{\textwidth}{\footnotesize *\SI{5}{\meter} RPE due to shorter
total distance traveled.}
\label{tab:results}
\vspace{-7mm}
\end{table}

\subsubsection{NCD-Multi Stairs} This sequence included
narrow spaces, mirrors, door opening and large rotations as shown in
Fig.~\ref{fig:challenges}. As a result, the baseline stereo visual odometry
struggled and lost track of all features several times.

\begin{figure}[t]
\centering
\vspace{2mm}
\includegraphics{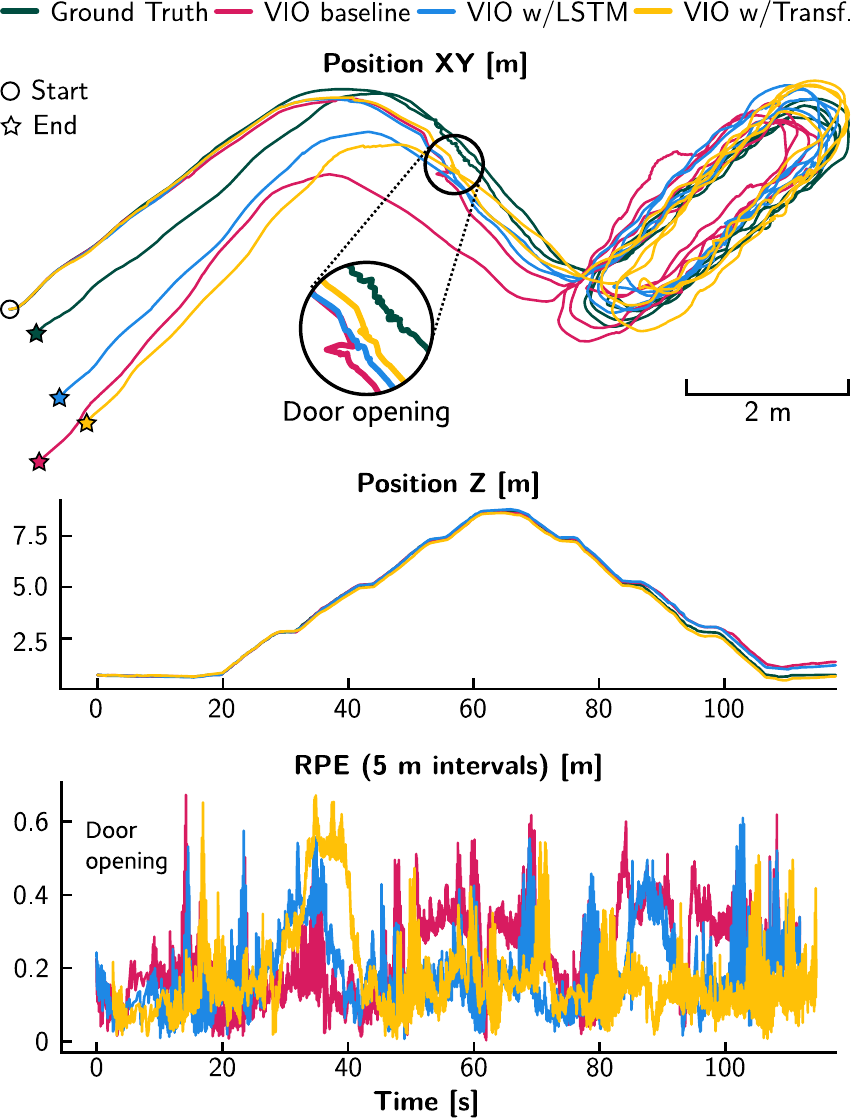}
\caption{NCD-Multi Stairs results. \textbf{Top: } Position
of the baseline VIO, LSTM and Transformer methods on $xy$-plane. We magnify the
door opening section of the trajectory where the baseline clearly diverges due
to loss of visual feature tracking and poor bias estimation. \textbf{Middle: }
Position of the trajectory for the $z$-axis. \textbf{Bottom: } \SI{5}{\meter}
RPE for each algorithm.}
\label{fig:stairs-plots}
\vspace{-5mm}
\end{figure}

We show in Fig.~\ref{fig:stairs-plots} that the learned bias predictions reduced relative pose errors. This is most clearly seen in the magnified
area where the door opening caused a significant drop in tracked visual
features. Our proposed approach improves upon pure IMU-only mechanization in
these situations where visual feature tracking fails.

\subsubsection{NCD-Multi Mine} The Mine sequence  was recorded in an abandoned
limestone mine in Corsham, UK. It was dark and dusty and the handheld platform
was intentionally held up very close to walls as shown in
Fig.~\ref{fig:challenges} on several occasions. This was also the longest
trajectory.

\begin{figure}[t]
\centering
\vspace{2mm}
\includegraphics{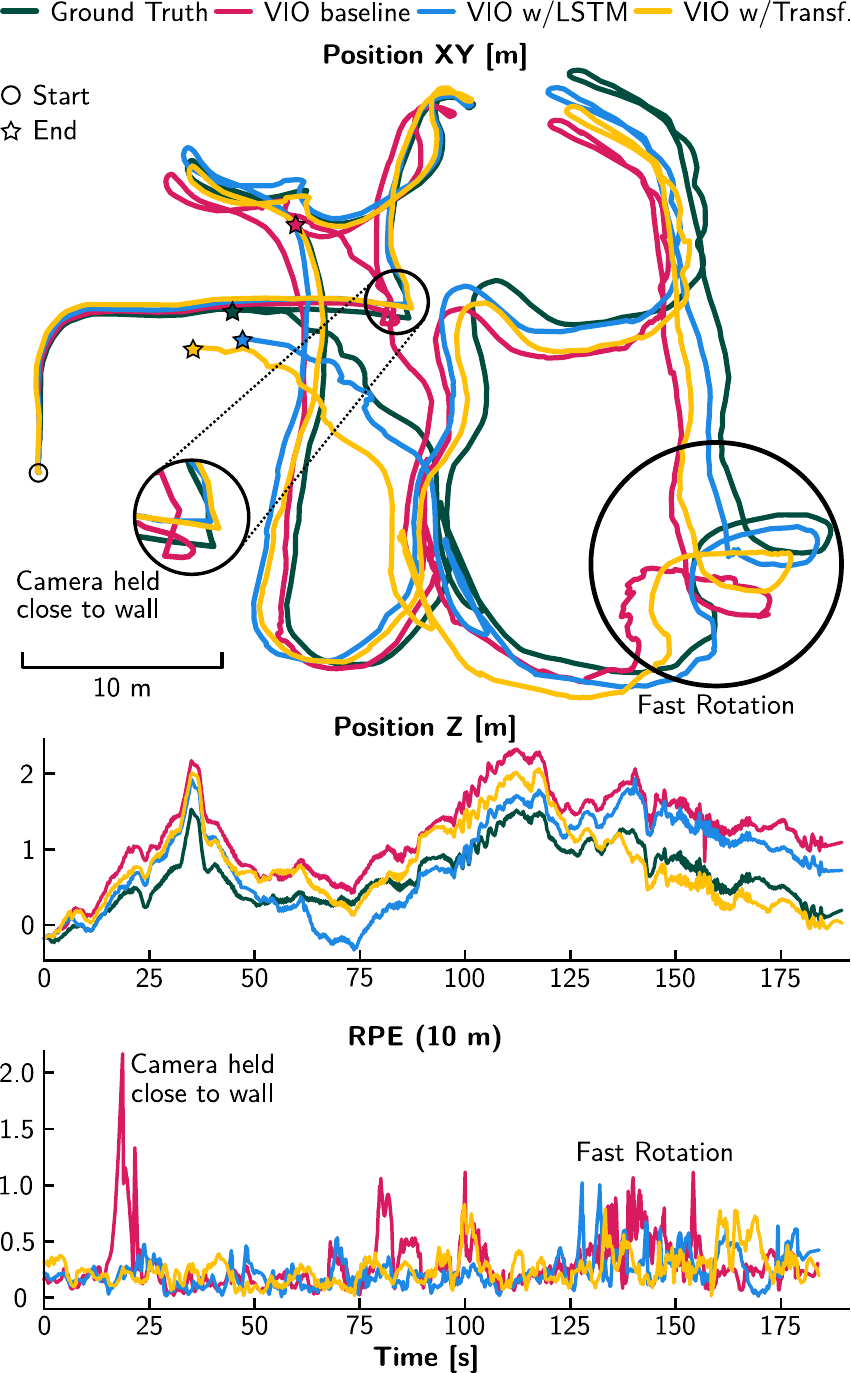}
\caption{Results from the Mine sequence. \textbf{Top: }
Position on $xy$-plane of the baseline VIO, LSTM and Transformer methods. We
magnify a part of the sequence where the camera was held facing and very close to a wall. The baseline diverges due to loss of visual feature tracking and poor bias estimation. We also highlight a section with large rotations where the baseline has multiple divergences. \textbf{Middle: } Position on $z$-axis. \textbf{Bottom: } \SI{10}{\meter} RPE for each algorithm.}
\label{fig:mine-plots}
\vspace{-5mm}
\end{figure}

The results can be seen in Fig.~\ref{fig:mine-plots}, where we highlight two
particular sections in which the baseline VIO became unstable. In one section the camera faced a wall for several seconds and another involved large
rotations in the dark. We see that by adding the bias prediction, trajectories are smoother and have lower error overall.

\subsubsection{NCD-Multi Quad} This sequence is long and in a
large-scale open space and includes several dynamic motions when the handheld
platform was shaken. Our method reduces error compared to the baseline. A summary of these results and for the other NCD-Multi sequences is provided in Table~\ref{tab:results}.

\subsection{Quadruped Robot Experiments}
\label{sec:quadruped}

For the Spot dataset, the same handheld sensor used in the NCD-Multi
dataset was mounted on a Boston Dynamics Spot (See Fig.~\ref{fig:platforms}) which was then
teleoperated around a courtyard in two separate
trajectories. The Easy trajectory is entirely on flat ground in good lighting
conditions while the Hard trajectory includes transitions from well-lit to shady
areas and walking on small ramps.

For these experiments, both the LSTM and Transformer were trained \textit{using only handheld data}. There was no fine-tuning for the quadrupedal robot. The
information used for training and testing was therefore exactly the same as in
Fig.~\ref{fig:motivation}. Numerical results are provided in
Table~\ref{tab:results}. This demonstrates that our method is independent from the locomotion modality and that the model can generalize across different modalities.

\subsection{Drone Experiments}
\label{sec:drone}
Two similar works to ours are Zhang et al.~\cite{zhang2021imu} and Brosard et
al~\cite{Brossard20ral}. Zhang et al. use an LSTM to denoise accelerometer and gyroscope measurements which are then passed to their visual-inertial pipeline based on~\cite{Mingming2021}. Brossard et al. use a CNN to denoise gyroscope measurements only which are then passed to Open-VINS~\cite{Geneva2020ICRA}. Since both works present their results using the EuRoC dataset~\cite{Burri25012016}, we also use this dataset to demonstrate that our method can be applied to drones. 
Because the baselines were evaluated differently, drawing conclusions from direct comparisons is difficult. Therefore, we also show relative improvement for each method.

Tables~\ref{tab:euroc-pos} and \ref{tab:euroc-ori} summarize these results in terms of absolute error. Note that \cite{zhang2021imu} reported positional RMSE while \cite{Brossard20ral} reported orientation RMSE and therefore we report both. For the V101 sequence, we use the re-computed ground truth from~\cite{Geneva2020ICRA} as did Brossard et al.~\cite{Brossard20ral}.

Brossard et al. additionally reported relative orientation errors by averaging
\SI{7}{\meter}, \SI{21}{\meter} and \SI{32}{\meter} relative orientation error
for all trajectories. Averaged over the three distances they report
\SI{1.59}{\degree} while we achieve \SI{1.41}{\degree} and \SI{1.36}{\degree}
for LSTM and Transformer respectively.

\subsection{Illustrative Experiments}
\label{sec:illustrative}
We present two additional experiments to further analyze our system's performance. In both cases we used the Spot Easy trajectory and models trained with only handheld data.

\subsubsection{Blackout}
\begin{figure}[t]
\centering
\vspace{2mm}
\includegraphics{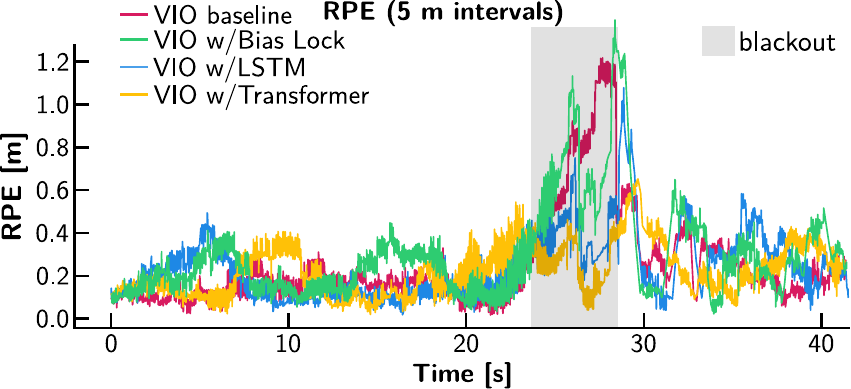}
\caption{\textbf{Blackout Experiment:} \SI{5}{\meter} RPE Performance where the
cameras were disabled for \SI{5}{\second} (gray area) during the Spot Easy
trajectory. Mean \SI{5}{\meter} RPE was \SI{0.26}{\meter}, \SI{0.28}{\meter}, \SI{0.22}{\meter} and \SI{0.20}{\meter} for the baseline, bias-lock, LSTM and Transformer methods respectively.}
\label{fig:discussion}
\vspace{-2mm}
\end{figure}

We ``black out" the images for \SI{5}{\second},
setting all pixels to $0$. As shown in Fig.~\ref{fig:discussion} there is an
error spike during the blackout where the estimator must rely on IMU data only.
We also experiment with ``locking'' the IMU biases during the blackout rather than
allowing the random walk.

\subsubsection{Image Distortion}
\begin{figure}[t]
\centering
\includegraphics[width=\linewidth]{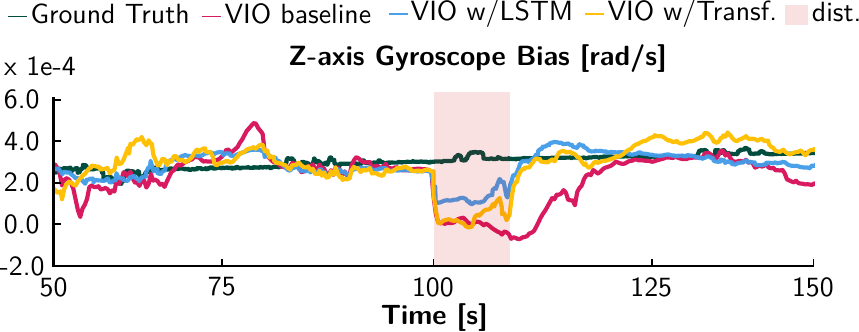}
\caption{\textbf{Image Distortion Experiment:} At \SI{100}{\second} of the Spot
Easy trajectory we apply a small distortion to both stereo images (pink area).
The baseline VIO develops a significant offset in $z$ bias as the error is attributed to the
gyroscope. LSTM and Transformer methods have reduced magnitude of error and recovery time.}
\label{fig:discussion2}
\vspace{-5mm}
\end{figure}

When there are errors in visual tracking, such as poor camera calibration or
wrong correspondences, the associated error can be erroneously attributed to
the IMU bias. Learned bias estimates can mitigate this by providing estimates of
the correct bias. To demonstrate this, we intentionally inject visual
tracking error by left shifting both stereo images by $10$ pixels for
\SI{10}{\second}. This is interpreted as an apparent rotation on the $z$-axis,
attributed by the system to a change in yaw gyro bias. In
Fig.~\ref{fig:discussion2} we plot the bias estimates during
this time. Even though we initialize all methods with the ground truth bias and use a robust loss function, each method's estimate of the gyroscope $z$ bias incorrectly drifts during the period of image distortion.

\begin{table}[t]
\centering
\vspace{2mm}
\caption{EuRoC: Absolute Position Error \& Improvement [\si{\meter} (\%)]}
\resizebox{\linewidth}{!}{
    \begin{tabular}{l|cc|ccc}  
    \toprule
     & \multicolumn{2}{c|}{\textbf{Zhang et al.~\cite{zhang2021imu}} } & \multicolumn{3}{c}{\textbf{Ours}}\\
    \textbf{Seq.} & \textbf{Baseline} & \textbf{Method} &
 \textbf{Baseline} &
\textbf{LSTM} &  \textbf{Transformer} \\
    \midrule
    MH02 & 0.19 & 0.15 (21\%) & 0.37 & 0.23 (38\%) & 0.13 \textbf{(65\%)} \\
    MH04 & 0.15 & 0.14 (7\%) & 0.33 & 0.25 \textbf{(24\%)} & 0.25 \textbf{(24\%)} \\
    V101 & 0.08 & 0.07 (13\%) & 0.17 & 0.08 \textbf{(53\%)} & 0.08 \textbf{(53\%)} \\
    V103 & 0.27 & 0.15 \textbf{(44\%)} & 0.27 & 0.27 (0\%) & 0.17 (37\%) \\
    V202 & 0.11 & 0.10 (9\%) & 0.29 & 0.23 (21\%) & 0.10 \textbf{(66\%)} \\
    \bottomrule
    \end{tabular}
    }
\label{tab:euroc-pos}
\end{table}

\begin{table}[t]
\centering
\caption{EuRoC: Absolute Rotation Error \& Improvement [\si{deg} (\%)]}
\vspace{-2mm}
\resizebox{\linewidth}{!}{
    \begin{tabular}{l|cc|ccc}  
    \toprule
     & \multicolumn{2}{c|}{\textbf{Brossard et al.~\cite{Brossard20ral}} } & \multicolumn{3}{c}{\textbf{Ours}}\\
    \textbf{Seq.} & \textbf{Baseline} &
\textbf{Method} & \textbf{Baseline} &
\textbf{LSTM} &  \textbf{Transformer}\\
    \midrule
    MH02 & 1.11 & 1.21 (-9\%) & 3.21 & 3.10 (3\%) & 2.86 \textbf{(11\%)} \\
    MH04 & 1.60 & 1.40 (13\%) & 0.89 & 1.00 (-12\%) & 0.76 \textbf{(15\%)} \\
    V101 & 0.80 & 0.80 (0\%) & 2.56 & 1.17 \textbf{(54\%)} & 1.66 (35\%) \\
    V103 & 2.32 & 2.25 (3\%) & 4.78 & 3.54 (26\%) & 1.87 \textbf{(61\%)} \\
    V202 & 1.85 & 1.81 (2\%) & 3.78 & 2.53 (33\%) & 1.31 (\textbf{65\%)} \\
    \bottomrule
    \end{tabular}
    }
  
\label{tab:euroc-ori}
\vspace{-5mm}
\end{table}

\section{Discussion}

In Section~\ref{sec:handheld} we show our methods achieving reduced error compared to the baseline in several visually challenging situations. On average the LSTM and Transformer have reduced error in Table~\ref{tab:results} by 15\%, but in certain situations, for example the camera held close to the wall in the Mine trajectory, the improvement is over 300\%. This is both because the learned estimates improve bias estimation and because they stabilize the optimizer in situations where error is incorrectly attributed to the IMU bias.

In Section~\ref{sec:quadruped} we take a model trained with handheld data and
apply it to quadrupedal data to demonstrate our method is locomotion agnostic.
Additionally, it should be noted the Stairs trajectory is the only handheld
trajectory with stair climbing and therefore that experiment also demonstrates
locomotion agnosticism. The reduced error of our method compared to the baseline shows our method is sensor specific and not locomotion specific.

In the Section~\ref{sec:drone} drone experiments, our method has similar accuracy as other methods. However, each method uses a different VIO baseline. Our method improves upon our baseline VIO more so than other methods improve on their baselines. Also, as the EuRoC dataset provides ground truth bias estimates, we examine the accelerometer and gyroscope bias $(\hba, \hbg)$ errors and find that, on average, error is reduced by (35\%,23\%) for LSTM and (32\%,12\%) for Transformer.

In the illustrative experiments in Section~\ref{sec:illustrative} we first present the Blackout experiment where our proposed methods reduced error during IMU-only estimation when no visual features could be tracked. Locking the bias did not improve on the baseline. This is because the bias random walk is already sufficiently constrained due to calibration. There is a second spike in error for all methods after vision is restored. This is because the pose estimate is suddenly corrected by a large amount, which locally appears as a relative error spike.

In the Image Distortion experiment, a bias shift affects all methods but the learned methods experience less error over the \SI{10}{\second} and a faster recovery period. During distortion, the error of LSTM and Transformer methods are 41\% and 12\% lower (RMSE) than the baseline.

LSTM and Transformer provide similar improvements to positional error with Transformer slightly better at rotational error. This makes sense because the gyroscope biases change more slowly and we expect the
Transformer to be more accurate with longer-term dependencies. In the distortion experiment the Transformer instead performs more poorly than the LSTM which we believe is due to the LSTM operating at \SI{2}{\hertz} allowing it to react to disturbances more quickly. In terms of model size the LSTM is superior with learned models as small as \SI{30}{\mega\byte} compared to \SI{200}{\mega\byte} for the transformer.
Additionally, the LSTM forward pass takes less than \SI{6}{\milli\second} as compared to \SI{20}{\milli\second} for the Transformer.

\section{Conclusion}
We present a novel application of machine learning to inertial navigation which
is more interpretable than similar methods. By learning bias predictions the
proposed method is more generally applicable because it is not robot specific and does not require periodic locomotion which we demonstrate with a wide variety of experiments on quadrupeds, handheld sensors and drones. We show that our method reduces \SI{5}{\meter} RPE of our baseline VIO system by 15\% on average in handheld and quadrupedal experiments with RPE being reduced by as much as 300\% in certain situations where vision fails. In the future we will build on this work by integrating uncertainty estimation into the model and experimenting with non-patterned locomotion such as wheeled robots.

\section*{ACKNOWLEDGMENT}
This work was supported by the EPSRC ORCA Robotics Hub (EP/R026173/1), the EU H2020 Project THING (Grant ID 780883) and a Royal Society University Research Fellowship (Fallon). For the purpose of Open Access, the author has applied a CC BY public copyright licence to any Author Accepted Manuscript (AAM) version arising from this submission.

\bibliographystyle{IEEEtran}
\bibliography{refs.bib}

\end{document}